\documentclass[letterpaper, 10 pt, conference]{ieeeconf}  % Comment this line out if you need a4paper

\IEEEoverridecommandlockouts                              % This command is only needed if 
\usepackage{url}
\title{\LARGE \bf
Knowledge Integration Strategies in Autonomous Vehicle Prediction and Planning: A Comprehensive Survey
}

\author{Kumar Manas and Adrian Paschke% <-this % stops a space
\thanks{K. Manas is with the Freie Universit{\"a}t Berlin, Germany and Continental Automotive Technologies GmbH, Germany. {\tt\small kumar.manas@fu-berlin.de}.\newline
A. Paschke is with the Freie Universit{\"a}t Berlin, Germany and Fraunhofer FOKUS Berlin, Germany.\newline
         This work is partially funded by the German Federal Ministry for Economic Affairs and Energy within the project ``nxtAIM".}% <-this % stops a space
}

\usepackage{textcomp}
\usepackage[absolute,showboxes]{textpos}
\TPMargin{5pt}

\newcommand{\copyrightstatement}{
    \begin{textblock}{0.82}(0.09,0.93)
         \noindent{\footnotesize{\copyright 2025 IEEE.
         Personal use of this material is permitted.
         Permission from IEEE must be obtained for all other uses, in any current or future media, including reprinting/republishing this material for advertising or promotional purposes, creating new collective works, for resale or redistribution to servers or lists, or reuse of any copyrighted component of this work in other works.

         \vspace{2mm}
         \noindent
         Accepted for publication in Proceedings of the IEEE Intelligent Vehicles Symposium (IV),  Cluj-Napoca - Romania, 22-25 June 2025.}}
    \end{textblock}
}   

\begin{document}

\maketitle
\copyrightstatement

%\begin{document}

\maketitle
\thispagestyle{empty}
\pagestyle{empty}

%%%%%%%%%%%%%%%%%%%%%%%%%%%%%%%%%%%%%%%%%%%%%%%%%%%%%%%%%%%%%%%%%%%%%%%%%%%%%%%%
\begin{abstract}

This comprehensive survey examines the integration of knowledge-based approaches in autonomous driving systems, specifically focusing on trajectory prediction and planning. We extensively analyze various methodologies for incorporating domain knowledge, traffic rules, and commonsense reasoning into autonomous driving systems. The survey categorizes and analyzes approaches based on their knowledge representation and integration methods, ranging from purely symbolic to hybrid neuro-symbolic architectures. We examine recent developments in logic programming, foundation models for knowledge representation, reinforcement learning frameworks, and other emerging technologies incorporating domain knowledge. This work systematically reviews recent approaches, identifying key challenges, opportunities, and future research directions in knowledge-enhanced autonomous driving systems. Our analysis reveals emerging trends in the field, including the increasing importance of interpretable AI, the role of formal verification in safety-critical systems, and the potential of hybrid approaches that combine traditional knowledge representation with modern machine learning techniques.

\end{abstract}

%%%%%%%%%%%%%%%%%%%%%%%%%%%%%%%%%%%%%%%%%%%%%%%%%%%%%%%%%%%%%%%%%%%%%%%%%%%%%%%%

\section{Introduction}
In modular automated driving, accurately predicting the maneuvers of traffic participants and planning safe driving actions are crucial. While significant progress has been made in simulation environments, the challenge lies in transitioning automated vehicles to the unpredictable real world, where they have to handle unforeseen behaviors and novel situations.
Rule-based systems perform well in structured environments but struggle with dynamic traffic conditions. Machine learning (ML)- based approaches typically require large datasets but struggles in out-of-distribution scenarios. In contrast, human drivers leverage a combination of rules, knowledge, and intuition to make informed decisions. Integrating such human-like reasoning—infusing rules and domain knowledge—into autonomous systems could enhance accuracy, robustness, and adaptability to unexpected scenarios.
Two key components discussed are \textbf{trajectory prediction}, which uses historical data and context such as road layout, agent interactions, and traffic rules to forecast the future paths of road users. Second, \textbf{trajectory planning} determines an optimal and feasible path for the ego vehicle, ensuring it complies with safety, legal, and comfort standards while anticipating the predicted movements of others.
Despite other surveys addressing various aspects of autonomous driving (AD), a notable gap exists in comprehensive reviews focusing specifically on integrating explicit knowledge representations into trajectory prediction and planning. Notably lacking are examinations of recent foundation and diffusion model-based methodologies, which have gained traction due to their ability to use vast amounts of unstructured data and advanced generative techniques to produce precise and adaptable predictions. In this paper, \textit{knowledge} is explicit representations of rules, ontologies, priors, textual information, or symbolic constraints, and \textit{symbolic approaches} are methods based on logic, rules, and ontologies without learned components.
Key observations driving this survey are:
\vspace{-1.5pt}
\begin{itemize}
\item \textbf{Fragmented Literature:} Previous reviews have explored trajectory prediction and planning separately \cite{surevey_predictor_no_KR}. Our survey uniquely examines how knowledge is holistically integrated across the automated driving domain, combining traditional methods with modern generative techniques.
\item \textbf{Emerging Methodologies:} Recent advances in foundation models, e.g., large language models (LLMs) and retrieval-augmented generation (RAG) and diffusion models offer promising avenues for integrating explicit regulatory and spatial constraints into prediction and planning, potentially revolutionizing how autonomous systems generate compliant, safe trajectories.

\item \textbf{Practical Relevance:} For real-world deployment, autonomous vehicles must operate accurately, adhere to traffic rules, and adapt to complex environments. Knowledge integration significantly enhances system robustness and safety, whether implicitly through data-driven models or explicitly via hybrid symbolic frameworks.
\end{itemize}
We focus exclusively on works that explicitly integrate rules or knowledge, excluding approaches that utilize environmental information without incorporating traffic regulations. However, in many cases, particularly in ML-based methods, this explicit prior knowledge does not become explicit; instead, in the overall pipeline, it becomes implicit or soft constraints. This survey provides researchers and practitioners with a unified perspective on achieving precise, rule-compliant trajectory prediction and planning by offering a detailed taxonomy and critical analysis of methods spanning classical approaches to the latest foundation and diffusion model-based techniques.
Our evaluation of symbolic and hybrid methodologies offers insights for developing next-generation autonomous systems that can understand and predict dynamic traffic scenarios while remaining safe, interpretable, and aligned with regulatory requirements.
\subsection{Related Surveys}
Several recent surveys have explored various aspects of AD, but none have specifically addressed the integration of domain knowledge into both prediction and planning modules. \cite{hagedorn2023review} reviews architectural principles for joint prediction and planning in deep learning systems but omits explicit rule embedding. \cite{chen2023end} analyzes end-to-end frameworks mapping raw sensor data to control commands yet excludes hybrid symbolic components. \cite{mehdipour2023formal} surveys formal verification techniques for rule compliance, focusing primarily on offline verification rather than runtime knowledge embedding. \cite{kgav2024integrating} reviews knowledge graph integration in AV systems, primarily at the perception layer. \cite{guo2023autonomous} provides a broad overview of AV system components without a detailed taxonomy of knowledge integration approaches.
In contrast, our survey delivers a unified taxonomy of knowledge integration methods spanning purely symbolic, hybrid neuro-symbolic, LLM-based, and diffusion-driven techniques, with comparative analysis of how traffic rules and domain priors are explicitly and implicitly encoded within both trajectory prediction and planning modules.

\begin{table*}[ht]
\centering
\caption{Overview of Knowledge and Traffic Rules Integration Methodology in Trajectory Prediction and Planning}
\label{tab:overview_all}
\vspace{-0.1cm}
\scriptsize
\begin{tabular}{p{1.8cm} p{3.1cm} p{5.9cm} p{5.1cm}}
\hline
\textbf{Method Category} & \textbf{Representative Works} & \textbf{Key Contributions \& Approach} & \textbf{Advantages / Limitations} \\
\hline
\textbf{Knowledge-Graph and Ontology-Based} &
\begin{itemize}
    \item \cite{Geng_Liang_Yu_Zhao_He_Huang_2017,zhao_intersection_ontology_2015,Balakirsky2004KnowledgeRA,Sun_Wang_Halilaj_Luettin_2024,Regele_2008,Jr_Jurmain_Ragazzi_2024,Fang_2019_ontology_reasoning_ITSC,zhang2022uti,Huang_2019_ontology,onto_crtical_westhofen,R_Uma_2018_ontology_survey,Giret_Julian_Carrascosa_Rebollo_2018,Bhuyan_2024_Ontology_Development_for_Sustainable_Intelligent,Kommineni_König-Ries_Samuel}
\end{itemize} &
\begin{itemize}
    \item Formalizes road elements, interactions, and regulatory constraints via OWL, ontologies, and KGs.
    \item Some works integrate probabilistic reasoning and hierarchical real-time decision–making.
\end{itemize} &
\begin{itemize}
    \item \textbf{Advantages:} High interpretability and explicit rule encoding.
    \item \textbf{Limitations:} Scalability issues and extensive manual engineering.
\end{itemize} \\
\hline
\textbf{Reinforcement Learning (RL)} &
\begin{itemize}
    \item \cite{Peiss_Wohlgemuth_Xue_Meyer_Gressenbuch_Althoff_2023,Lin_Zhou_Wang_Cao_Yu_Zhao_Zhao_Yang_Li_2022,Yuan_2024_Evolutionary_Decision_Making_and_Planning,Li2024Let,tang_2023_Personalized_Decision_Making_and_Control,pan2017virtual,kiran_Deep_Reinforcement_Learning,cusumano2025robust}
\end{itemize} &
\begin{itemize}
    \item Embeds traffic rules in state representations and reward functions.
    \item Employs hybrid strategies (e.g., imitation learning, formal logic) to improve safety and exploration.
\end{itemize} &
\begin{itemize}
    \item \textbf{Advantages:} Adaptive policy learning from dynamic environments.
    \item \textbf{Limitations:} Reward engineering complexity, extensive simulation, and sample inefficiency.
\end{itemize} \\
\hline
\textbf{LLM \& RAG Methods} &
\begin{itemize}
    \item \cite{Li_2024_CVPR,Cai_Liu_Zhou_Ma_Zhao_Wu_Ma_2024,tr2mtl,xu2024drivegpt4,Mao2023ALA,hwang2024emmaendtoendmultimodalmodel,Cui2024DriveLLM:,sha2023languagempclargelanguagemodels,Fu_Li_Wen_Dou_Cai_Shi_Qiao_2024,Yuan_Sun_Omeiza_Zhao_Newman_Kunze_Gadd,LLMAssistClosedLoopPlanning,wen2024diluknowledgedrivenapproachautonomous}
\end{itemize} &
\begin{itemize}
    \item Uses unstructured text and regulatory documents to adapt driving behavior.
    \item Translates natural language rules into formal (temporal) logic and generates context-aware decisions.
    \item Often combined with multimodal inputs (e.g., DriveGPT4).
\end{itemize} &
\begin{itemize}
    \item \textbf{Advantages:} Dynamic adaptation and enhanced interpretability.
    \item \textbf{Limitations:} Latency concerns and risk of LLM hallucinations.
\end{itemize} \\
\hline
\textbf{Formal Logic-Based Methods} &
\begin{itemize}
    \item \cite{Karimi_Duggirala_2020,auto_discern,suchan2023assessing,Manzinger2021Using,Koschi2021Set-Based,Loos_Platzer_Nistor_2011,Bhuiyan_Governatori_Bond_Demmel_Badiul_2020,Chan_Li_Lu_Lin_Bundy_2025,esterle2020,coogan2014,raman2017,Sadraddini2016,Lin2022ModelPR}
\end{itemize} &
\begin{itemize}
    \item Formalizes traffic regulations via temporal logic, ASP, etc.
    \item Integrated into verification, MPC, and optimization frameworks for provable safety.
\end{itemize} &
\begin{itemize}
    \item \textbf{Advantages:} Strong safety guarantees and high interpretability.
    \item \textbf{Limitations:} Worst-case planning time grows exponentially with rule size and challenges with real-time integration.
\end{itemize} \\
\hline
\textbf{Hybrid Symbolic \& Neural Methods} &
\begin{itemize}
    \item \cite{auto_discern} (also in Formal Logic),
    \item \cite{Bahari_Nejjar_Alahi_2021,Li_Rosman_Gilitschenski_DeCastro_Vasile_Karaman_Rus_2021,li2023towards,sormoli2024survey,badreddine2022logic,manhaeve2018deepproblog,manas2025shift,kamale_Cautious_Planning_with_Incremental,Li2021Vehicle,phan2022driving,vitelli2022safetynet,patrikar2024rulefuser}
    \item \cite{Fu_Li_Wen_Dou_Cai_Shi_Qiao_2024,wen2024diluknowledgedrivenapproachautonomous} (also in LLM)
  
\end{itemize} &
\begin{itemize}
    \item Fuses deep learning–based perception with explicit symbolic reasoning.
    \item Incorporates symbolic constraints in loss functions or via LLM–generated rules.
\end{itemize} &
\begin{itemize}
    \item \textbf{Advantages:} Balances flexibility with rule compliance.
    \item \textbf{Limitations:} Increased integration complexity and synchronization challenges.
\end{itemize} \\
\hline
\textbf{Diffusion \& Other Learning-Based Approaches} &
\begin{itemize}
    \item \cite{patrikar2024rulefuser,stl_diffusion,CoBL_diffusion}
\end{itemize} &
\begin{itemize}
    \item Leverages diffusion processes to generate multimodal, rule–compliant trajectories.
    \item Blends stochastic exploration with explicit constraint encoding.
\end{itemize} &
\begin{itemize}
    \item \textbf{Advantages:} Captures diverse trajectory distributions.
    \item \textbf{Limitations:} Balancing exploration and speed with strict rule adherence can be challenging.
\end{itemize} \\
\hline
\end{tabular}
\vspace{-0.7cm}
\end{table*}

\section{Literature Survey}
This section presents a high-level overview of the diverse methodologies for integrating domain knowledge, traffic rules, and commonsense reasoning into AD systems. Our survey is organized into several subsections—ranging from knowledge graph and ontology-based methods, reinforcement learning (RL) methods, and LLM approaches to formal logic-based and hybrid neuro-symbolic methods—that collectively cover the spectrum of current research. To navigate the diverse methodologies, Table~\ref{tab:overview_all} offers a concise, comparative summary of the key studies covered in this survey. The table categorizes each representative study by its methodological approach, emphasizes the main contributions, and outlines strengths and weaknesses. Some works fall under multiple categories; we assign each a primary class but list cross-references. This high-level overview facilitates understanding how these diverse approaches interconnect and serves as a roadmap for the detailed discussions in subsequent sections.

\subsection{Knowledge-Graph and Ontology-Based Methods}
\label{ontology_and_KG}
Recent advancements in AD have highlighted the importance of structured knowledge representation using ontologies and knowledge graphs (KGs) to model complex urban environments, enhance prediction accuracy, and enable rule-compliant planning. These frameworks provide explicit semantic representations of static road elements (e.g., lanes, traffic signs), dynamic interactions (e.g., vehicle-pedestrian relationships), and regulatory constraints, enabling interpretable reasoning across heterogeneous data sources.

Early works demonstrated the potential of ontology-based reasoning for driving scenarios. \cite{Geng_Liang_Yu_Zhao_He_Huang_2017} pioneered a scenario-adaptive approach using Web Ontology Language (OWL) to formalize spatial states and vehicle-road relationships, integrating these with hidden Markov models for maneuver prediction. At the same time, their method improves prediction horizons through domain knowledge, and manual ontology engineering limits scalability. Similarly, \cite{zhao_intersection_ontology_2015} developed intersection-specific ontologies with rule-based decision-making, and \cite{Balakirsky2004KnowledgeRA} established a framework for querying static and dynamic traffic semantics in real time. These studies underscore the trade-off between interpretability and adaptability: rule-based ontologies excel in structured environments like intersections \cite{zhao_intersection_ontology_2015} but struggle with unseen scenarios.

SemanticFormer \cite{Sun_Wang_Halilaj_Luettin_2024} improves trajectory prediction by considering traffic participants, road topology, and signs, enhancing performance by 4-5\% when added to existing trajectory predictors. Hybrid methods combining ontologies with probabilistic models address uncertainty inherent in traffic interactions. \cite{Regele_2008} merged high-level traffic coordination ontologies with low-level trajectory models, while \cite{Jr_Jurmain_Ragazzi_2024} integrated description logic with probabilistic semantics to handle incomplete information. \cite{Fang_2019_ontology_reasoning_ITSC} extended this by encoding interaction probabilities for long-term behavior prediction, though computational delays hindered real-time control. \cite{zhang2022uti} augmented scene understanding by integrating KGs extracted from textual corpora, enabling causal reasoning for novel scenarios. However, offline knowledge curation in their approach limits responsiveness to dynamic environments.

Ontologies also support real-time decision-making through hierarchical reasoning. \cite{Huang_2019_ontology} leveraged driving experience knowledge bases for situation estimation, while \cite{onto_crtical_westhofen} modeled critical factors via joint description logic and rule-based reasoners. These methods enhance transparency in regulatory compliance but face scalability challenges as ontology complexity grows \cite{Balakirsky2004KnowledgeRA}. A comprehensive survey by \cite{R_Uma_2018_ontology_survey} emphasizes that while ontological frameworks improve interoperability and rule awareness, their rigidity complicates adaptation to unexpected events.

Current research focuses on mitigating manual engineering burdens through standardization and automation. Domain-specific ontologies like ASAM OpenXOntology\footnote{\url{www.asam.net/standards/asam-openxontology}} provide reusable templates for traffic concepts, reducing development efforts \cite{Giret_Julian_Carrascosa_Rebollo_2018, Bhuyan_2024_Ontology_Development_for_Sustainable_Intelligent}. Promisingly, LLMs are being explored for semi-automated KG generation \cite{Kommineni_König-Ries_Samuel}, potentially enabling dynamic knowledge integration with sensor data. Such advances aim to balance the interpretability of symbolic reasoning with the adaptability of data-driven methods, addressing long-standing challenges in real-time performance and scalability \cite{R_Uma_2018_ontology_survey, Fang_2019_ontology_reasoning_ITSC}.

\subsection{Reinforcement Learning (RL) Methods}
\label{RL_approach}
Reinforcement learning (RL) has emerged as a powerful paradigm for developing AD systems, enabling sequential decision-making through dynamic environment interactions. By integrating safety constraints, regulatory compliance, and structured domain knowledge into reward functions and state representations, RL frameworks excel in learning adaptive policies for complex driving scenarios. Recent advancements in this domain focus on curriculum learning (progressively increasing the complexity of training scenarios), safety-centric reward design, and sim-to-real transfer while addressing challenges such as sample inefficiency and reward engineering.

A key trend in RL-based methods is using structured state encodings to embed traffic rules and safety priors. For instance, \cite{Peiss_Wohlgemuth_Xue_Meyer_Gressenbuch_Althoff_2023} proposes a graph-based RL framework that employs curriculum learning to progressively increase the complexity of highway driving scenarios. By integrating explicit traffic-rule information into the state representation using graph neural networks, their method accelerates learning and improves collision avoidance and lane-change efficiency in dense traffic. Similarly, \cite{Lin_Zhou_Wang_Cao_Yu_Zhao_Zhao_Yang_Li_2022} addresses the challenge of adapting to evolving traffic regulations by designing a law-adaptive backup policy. Their approach translates natural language traffic laws into formal constraints, enabling the RL agent to recognize potential violations and trigger re-planning. While these methods enhance compliance and safety, they introduce the complexity of reward engineering and policy evaluation.

Safety-centric RL frameworks often enforce rule adherence through reward shaping and hybrid architectures. \cite{Yuan_2024_Evolutionary_Decision_Making_and_Planning} augment the traditional RL reward function with explicit penalties for traffic violations, balancing goal achievement with safety metrics. This evolutionary RL framework emphasizes safe and rational exploration, guiding the agent toward policies that adhere to safety standards. In a similar vein, \cite{Li2024Let} combines deep RL with Hybrid A* for local trajectory planning, using linear temporal logic (LTL) to formalize traffic rule compliance in the reward function. Imitation learning hybrids, such as the generative adversarial imitation learning framework proposed by \cite{tang_2023_Personalized_Decision_Making_and_Control}, refine policies through adversarial feedback on expert demonstrations. These approaches yield human-like behaviors while ensuring rule compliance, though they require careful calibration to avoid over-constraining exploration.

Bridging the gap between simulation and real-world deployment remains a critical challenge for RL-based planners. \cite{pan2017virtual} investigates the sim-to-real transfer of RL policies, training agents in virtual environments and adapting them to real-world conditions. Their work highlights the potential of RL for robust performance in heterogeneous scenarios but also underscores the challenges of achieving reliable generalization. Surveys such as \cite{kiran_Deep_Reinforcement_Learning} provide a broader perspective, categorizing RL-based approaches by state representation (e.g., raw sensor inputs vs. graph abstractions) and reward formulations. These surveys emphasize the trade-offs between sample efficiency and safety guarantees, offering insights into the design of scalable and reliable RL frameworks.

Despite their advantages, RL-based methods face several challenges. Reward engineering remains a non-trivial task, as balancing safety, compliance, and efficiency in reward design is critical for policy performance. Sample inefficiency is another significant bottleneck, as complex environments demand extensive training data, prolonging development cycles. The sim-to-real gap further complicates deployment, with discrepancies between simulated and real-world dynamics hindering reliable policy transfer. Additionally, the ``black-box" nature of RL policies complicates verification in safety-critical contexts, necessitating interpretable and verifiable representations.

Emerging directions in the RL-based model focus on addressing these challenges. Evolutionary strategies, such as those proposed by \cite{Yuan_2024_Evolutionary_Decision_Making_and_Planning}, offer efficient exploration mechanisms for safe policy learning. Formal methods like linear temporal logic (LTL) \cite{Li2024Let} provide verifiable compliance guarantees, while hybrid architectures blending imitation learning with RL \cite{tang_2023_Personalized_Decision_Making_and_Control} enhance human-like behavior and adaptability. Recent work in Multi-Agent RL (MARL) using self-play shows promise for achieving robust performance and handling complex interactions in simulated real-world benchmarks \cite{cusumano2025robust}, potentially addressing some scalability and interaction modeling challenges. Future research should prioritize scalable reward automation, robust sim-to-real pipelines, and interpretable policy representations to advance the real-world applicability of RL-based autonomous driving systems.

\subsection{Large Language Model and Retrieval-Augmented Generation Methods}
\label{LLM_and_KG}
LLMs and RAG techniques have become valuable tools for enhancing prediction and planning systems. By automating the transformation of unstructured and regulatory data into intermediate representations, they improve decision-making, adaptability, and interpretability. These approaches enable systems to access vast knowledge bases dynamically, generate human-readable explanations, and adapt to region-specific traffic regulations, effectively addressing critical challenges in predictive and planning tasks.

A prominent application of LLMs is their ability to interpret and adapt to traffic regulations in real time. LLaDA (Large Language Driving Assistant)~\cite{Li_2024_CVPR} exemplifies this capability, using LLMs to adapt driving behavior to new environments, customs, and laws. By interpreting traffic rules from local driver handbooks, LLaDA provides context-aware driving instructions and demonstrates zero-shot generalization. Similarly,~\cite{Cai_Liu_Zhou_Ma_Zhao_Wu_Ma_2024} propose a framework where a Traffic Regulation Retrieval Agent employs RAG to query regional traffic documents and an LLM-based reasoning module evaluates maneuvers for compliance and safety. While this approach dynamically adapts to region-specific laws and generates rule-referenced explanations, it faces latency and LLM hallucination challenges. The TR2MTL framework \cite{tr2mtl} further advances this paradigm by using LLMs in a chain-of-thought learning paradigm to translate natural language traffic rules into metric temporal logic formulas, reducing manual rule specification and enabling scalable regulatory updates.

LLMs are increasingly integrated with multimodal data to enhance perception and decision-making. DriveGPT4 \cite{xu2024drivegpt4} represents an end-to-end system that generates driving decisions directly from sensory inputs, providing natural language explanations for its trajectory predictions. This approach improves interpretability and generalization in novel environments. Mao et al. \cite{Mao2023ALA} further contribute by developing a language agent that fuses textual reasoning with visual perception, retrieving relevant rule excerpts and synthesizing actionable driving strategies. Industry efforts, such as Waymo’s EMMA model \cite{hwang2024emmaendtoendmultimodalmodel}, highlight the potential of multimodal LLMs like Google’s Gemini to process sensor data and predict future trajectories, signaling a shift toward LLM-centric autonomous systems.

Recent works explore using LLMs as central decision-making components in AD. DriveLLM \cite{Cui2024DriveLLM:} charts a comprehensive roadmap for integrating LLMs into fully autonomous driving systems, emphasizing their role in high-level reasoning, planning, and interaction, while the DiLu framework \cite{wen2024diluknowledgedrivenapproachautonomous} specifically leverages LLM reasoning and reflection modules for knowledge-driven decision-making using commonsense and accumulated experience. Similarly, LanguageMPC \cite{sha2023languagempclargelanguagemodels} proposes a framework where LLMs serve as decision-makers within a model predictive control (MPC) pipeline. This approach combines the reasoning capabilities of LLMs with the robustness of traditional control methods, enabling the system to generate human-like driving strategies while ensuring safety and compliance. Both frameworks highlight the potential of LLMs to bridge the gap between high-level reasoning and low-level control. However, challenges such as real-time performance and integration with existing systems remain.

LLMs and RAG techniques are also employed to augment reasoning with commonsense and regulatory knowledge. \cite{zhang2022uti} utilize dense retrieval mechanisms and LLMs to answer domain-specific questions about traffic scenarios, implicitly integrating background knowledge into decision-making. The “Drive Like a Human” framework \cite{Fu_Li_Wen_Dou_Cai_Shi_Qiao_2024} enhances generalization and interpretability by combining LLM-generated high-level context descriptions with neural perception modules. RAG-Driver \cite{Yuan_Sun_Omeiza_Zhao_Newman_Kunze_Gadd} extends this concept by employing a multimodal LLM for driving explanations and control predictions, leveraging retrieval-augmented in-context learning to achieve strong zero-shot generalization. Additionally, LLM-Assist \cite{LLMAssistClosedLoopPlanning} demonstrates the synergistic potential of LLMs in hybrid planning architectures, using them alongside conventional rule-based planners to improve performance in complex scenarios.

Despite their promise, LLM and RAG-based methods face several challenges. Real-time performance remains a critical bottleneck, as the computational demands of LLMs can hinder their deployment in latency-sensitive applications. The risk of hallucinations—incorrect or nonsensical outputs—poses safety concerns, particularly in dynamic driving environments. Effective fusion of multimodal sensory data with language-based reasoning also requires further research to ensure robust and reliable decision-making. Ongoing efforts aim to address these issues through advancements in model efficiency, hallucination mitigation, and multimodal integration.

LLM and RAG-based methods offer a promising pathway to adaptive, transparent, and compliant autonomous driving systems. By dynamically incorporating regulatory knowledge, generating interpretable explanations, and enabling zero-shot generalization, these approaches address key limitations of traditional methods. However, realizing their full potential will require overcoming real-time performance, reliability, and multimodal fusion challenges, paving the way for next-generation autonomous systems capable of navigating complex and rapidly changing environments.

\subsection{Formal Logic-Based Methods}
\label{formal_logic_method}
Formal logic-based methods have become a cornerstone in autonomous driving research, offering rigorous frameworks for monitoring, verification, prediction, and planning through mathematically grounded representations of traffic rules and safety requirements. A central theme in this work is the formalization of traffic regulations using temporal logic and answer set programming (ASP). For instance, \cite{Karimi_Duggirala_2020} formalizes traffic rules at uncontrolled intersections by encoding yielding and right–of–way as temporal logic formulas, ensuring that vehicle behaviors adhere to safe bounds, even though extending this approach to complex urban scenarios remains challenging and \cite{esterle2020} formalized traffic regulations for dual carriageways in linear temporal logic (LTL), providing a machine-interpretable rule set. Similarly, AUTO-DISCERN \cite{auto_discern} employs ASP via the goal-directed s(CASP) system to simulate commonsense reasoning in driving decisions. By encoding driving rules as explicit logical constraints and generating proof trees to justify maneuvers, AUTO-DISCERN achieves high interpretability while grappling with scalability issues and sensor noise.

Another emerging direction involves integrating formal logic with data-driven techniques to enhance robustness in trajectory prediction. RuleFuser \cite{patrikar2024rulefuser} exemplifies this trend by combining neural predictors with classical rule-based predictors; here, formal logic is used to represent traffic rules that guide predictions in out–of–distribution scenarios. %In a related effort, the TR2MTL framework \cite{tr2mtl} leverages large language models (LLMs) to automatically translate unstructured natural language traffic rules into metric temporal logic (MTL) formulas. By employing a chain–of–thought in–context learning strategy, TR2MTL reduces manual rule formalization, although its performance is sensitive to prompt design and input quality.

Formal logic is also employed to enhance situational awareness and decision-making. \cite{suchan2023assessing} applies ASP–based reasoning to model and assess driver situation awareness in semi-autonomous vehicles, formalizing scene interpretation as logical rules that verify correct perception and reaction to dynamic traffic conditions. Further reinforcing the role of formal methods, safety verification techniques based on differential dynamic logic \cite{Loos_Platzer_Nistor_2011} have been applied to prove the safety of vehicle control systems under varied driving conditions. Such methods, along with survey works like \cite{mehdipour2023formal}, illustrate that while formal logic-based approaches provide strong guarantees and a high level of interpretability, they often face challenges in terms of computational complexity, scalability, and real-time integration of continuously changing sensor inputs.

Defeasible Deontic Logic (DDL) has been applied to formalize traffic rules, enabling autonomous vehicles to reason about obligations, permissions, and prohibitions in dynamic driving environments. By encoding traffic regulations using DDL \cite{Bhuiyan_Governatori_Bond_Demmel_Badiul_2020,Chan_Li_Lu_Lin_Bundy_2025}, vehicles can predict trajectories that comply with legal and ethical standards while also accommodating exceptions and resolving conflicts among rules. This approach enhances the interpretability and adaptability of trajectory prediction systems, allowing for more nuanced decision-making in complex traffic scenarios. However, implementing DDL in real-time applications poses challenges due to its computational complexity and the need for comprehensive rule databases. Balancing the expressiveness of DDL with the efficiency required for real-time trajectory prediction remains a critical area for further research.

In practice, many of these formal logic-based methods are integrated into hybrid and optimization–based models. Formal logic offers a versatile toolkit for ensuring provably safe and compliant autonomous driving behaviors, whether employed as constraints in loss functions, integrated into reinforcement learning reward structures, or used in optimization layers for planning. However, manual formalization is a tedious process. To solve this problem, TR2MTL~\cite{tr2mtl} attempts to automate this process using LLM and treating text to formal logic as a translation task. Despite their promise, ongoing research continues to address the trade-offs between the rigor of formal methods and the need for real-time, scalable performance in dynamic driving environments. 

\textbf{Formal Logic in MPC and Optimization.} Model Predictive Control (MPC) is a widely adopted optimization-based control strategy in autonomous driving. It solves an optimal control problem over a finite receding horizon to generate trajectories that account for both current sensor data and predicted future states. This approach translates traffic rules into mathematical inequalities or cost functions directly embedded into optimization-based planning frameworks. By formulating safety rules as constraints within MPC, recent research has increasingly focused on integrating formalized traffic rules into these optimization frameworks to enhance both performance and safety. Embedding traffic constraints as mathematical inequalities and temporal logic conditions allows MPC-based planners to enforce rigorous safety and regulatory requirements while optimizing for efficiency.
Another influential approach merges MPC with Signal Temporal Logic (STL) specifications. In \cite{raman2017}, STL constraints are encoded as mixed-integer linear constraints within an MPC framework. This formulation allows the controller to systematically enforce temporal properties during receding horizon control, which is particularly beneficial for applications requiring stringent timing and safety adherence. Extending these ideas to the urban context, \cite{Sadraddini2016} applies MPC to optimize traffic signal operations under temporal logic constraints, ensuring that predefined temporal conditions are met to reduce congestion and improve traffic flow. Building on the concept of robustness, \cite{Lin2022ModelPR} incorporates predictive robustness within an STL-based MPC framework, enhancing the system's reliability in the face of uncertainties in dynamic environments.

\textbf{Set‐Based Reachability Methods} compute over‐approximations of all possible future states (reachable sets)—to guarantee collision avoidance across all admissible maneuvers. \cite{Manzinger2021Using} combines convex over‐approximations of the ego vehicle’s reachable set with real‐time convex optimization, yielding faster planning as scenarios are constrained. \cite{Koschi2021Set-Based} extends this concept to prediction, formally enclosing visible and occluded agent behaviors via nondeterministic motion models and traffic‐rule constraints. These approaches complement temporal-logic MPC planners by providing provable safety envelopes over entire state tubes rather than single nominal trajectories.

Control frameworks for traffic networks, where signalized intersections are managed by synthesizing control policies that satisfy temporal logic constraints, are developed by \cite{coogan2014}. Their approach leverages formal methods and model checking to generate strategies that guarantee compliance with complex operational constraints, thereby improving safety and efficiency in urban traffic management.
Collectively, these studies illustrate a significant trend: the integration of formal traffic rules into optimization–based control strategies yields control systems that are both flexible and provably safe. The combined use of MPC and temporal logic—whether through LTL or STL—offers a promising pathway toward designing controllers capable of real-time adaptation while rigorously satisfying safety and regulatory requirements. This hybridization of formal logic and optimization advances autonomous driving and paves the way for future developments in intelligent transportation systems. For a comprehensive survey explicitly focused on formal methods for traffic rule compliance in autonomous driving, readers are referred to \cite{mehdipour2023formal}.

\subsection{Hybrid Approaches Combining Symbolic and Neural Methods}
\label{hybrid approach}

Hybrid approaches for autonomous driving have evolved by combining the adaptability of neural networks with the rigor and interpretability of symbolic reasoning. These methods leverage the pattern recognition power of deep learning while enforcing implicit and explicit rule-based constraints, enhancing safety, robustness, and explainability. These approaches can be broadly categorized by their integration strategy.

One notable method, AUTO-DISCERN \cite{auto_discern}, integrates deep neural perception pipelines with symbolic reasoning modules. They process sensor inputs through deep networks, whose features feed into an answer set programming-based module (using the goal-directed s(CASP) system) to apply commonsense rules and regulate driving behaviors.

A second category focuses on integrating symbolic knowledge directly into the learning process. In \cite{Bahari_Nejjar_Alahi_2021}, symbolic constraints are injected into the loss functions and network architectures, enhancing robustness in rare or hazardous scenarios. This idea is further developed in \cite{Li_Rosman_Gilitschenski_DeCastro_Vasile_Karaman_Rus_2021}, where formal rules expressed as temporal logic is incorporated in a differentiable manner, penalizing non–compliant predictions. In a related effort, \cite{Li2021Vehicle} employs signal temporal logic and syntax trees as features within a generative adversarial network framework to improve trajectory prediction accuracy without biasing the system away from lawful behavior. Similarly, the SHIFT \cite{manas2025shift} combines explicitly encoded traffic rules with learning–based uncertainty-aware trajectory predictors; here, traffic rules are used as a soft constraint and provide prior information to the predictor, guiding the neural network towards rule–compliant predictions while reducing the data requirements.

Complementing methods that inject explicit symbolic rules, another form of hybrid integration involves learning implicit behavioral rules or preferences directly from data. For example, approaches using Inverse Reinforcement Learning (IRL) infer complex reward functions from human driving demonstrations, capturing nuanced objectives that may implicitly include safety and comfort considerations, effectively learning the `knowledge' embedded in expert behavior \cite{phan2022driving}.

A recent trend leverages LLMs to bridge high–level symbolic reasoning and sensor-driven perception. For instance, \cite{Fu_Li_Wen_Dou_Cai_Shi_Qiao_2024} utilizes an LLM to generate high–level symbolic descriptions of the driving context, which are then fused with deep learning–based sensor interpretations. Building on this concept, DILU \cite{wen2024diluknowledgedrivenapproachautonomous} automatically generates symbolic rules from unstructured data via LLMs and integrates them into the decision–making pipeline. Although promising, these systems still face challenges such as LLM hallucinations, integration latency, and the synchronization of subsymbolic and symbolic processing streams.

Complementing these technical approaches, several surveys and alternative frameworks provide a broader perspective on the hybrid paradigm. Reviews in \cite{li2023towards} and the survey on hybrid motion planning \cite{sormoli2024survey} underscore the potential of embedding domain knowledge into learning–based systems to enhance safety and interpretability. Additionally, neural–symbolic methods like Logic Tensor Networks \cite{badreddine2022logic} and DeepProbLog \cite{manhaeve2018deepproblog} offer frameworks for integrating logical constraints into the learning-based predictors and planners.

Finally, hybrid architectures are designed to ensure safety and handle real-time demands. One common pattern involves a `safety net' \cite{vitelli2022safetynet} approach, where a primary machine-learned policy is overseen by a simpler, verifiable safety controller that intervenes if unsafe actions are proposed. Other specialized systems focus on real-time adaptability and conflict resolution. For example, \cite{manas_legal_2023} formalizes explicit traffic rules to rule out infeasible maneuvers, and \cite{kamale_Cautious_Planning_with_Incremental} combines incremental symbolic perception with reactive control synthesis to generate verified, rule–compliant driving maneuvers in dynamic environments.

In summary, hybrid approaches that integrate symbolic and neural methods offer significant advantages by combining explicit rule-based reasoning with the flexibility of deep learning or by structuring systems with complementary learned and verifiable components. While these methods enhance interpretability and safety in complex driving scenarios, challenges remain in managing computational complexity, ensuring real-time performance, and effectively synchronizing heterogeneous components.

\subsection{Diffusion and Other Learning-Based Prediction Incorporating Traffic Rules}

Diffusion-based models have recently shown promising results in generating rule-compliant trajectories while adhering to user-defined constraints. These models can encode traffic rules as STL within the diffusion process \cite{stl_diffusion} or integrate control barrier functions and Lupanov function-based reachability encoding \cite{CoBL_diffusion}. Diffusion models are particularly effective at capturing multimodal distributions and generating diverse trajectories while maintaining constraint satisfaction. However, they often require careful design to balance stochastic exploration with rule adherence.

Beyond purely formal and optimization-based approaches, learning-based trajectory prediction has increasingly incorporated explicit traffic rules and regulatory constraints. These methods enhance deep neural models by embedding formal rules into loss functions or input representations, ensuring predicted trajectories optimize performance while complying with safety and legal standards. For example, RuleFuser \cite{patrikar2024rulefuser} demonstrates that combining symbolic reasoning with data-driven predictors improves robustness in out-of-distribution scenarios. Similarly, approaches that integrate temporal logic constraints into the training process \cite{Li_Rosman_Gilitschenski_DeCastro_Vasile_Karaman_Rus_2021} yield dynamically feasible and rule-compliant trajectories.

\section{Discussion and Future Directions}  

This survey systematically reviews methodologies for trajectory prediction and planning in autonomous driving. We define trajectory prediction as forecasting agents’ future paths using historical and contextual data and trajectory planning as generating safe, collision-free paths for the ego vehicle under regulatory constraints. Our analysis spans classical methods (e.g., geometric models, optimization-based strategies) to modern hybrid frameworks merging symbolic reasoning with deep neural networks.  

A central focus lies in emerging foundation model- and diffusion model approaches, which leverage unstructured data and generative mechanisms to encode traffic dynamics either via explicit symbolic modules or implicitly within learned representations. These methods promise to enhance prediction accuracy and planning robustness in complex scenarios, signaling a paradigm shift in autonomous systems. Our review equips researchers with methodologies to address prediction-planning challenges while balancing safety, efficiency, and goal achievement.  

Future work can resolve real-time performance in dynamic environments, improve the interpretability of deep learning systems, and unify heterogeneous data streams (e.g., sensor inputs and traffic rules). The scalability and adaptability of foundation/diffusion models in diverse driving conditions require further validation. Multimodal reasoning and formal safety guarantees—particularly for handling nuanced, conflicting rules and hierarchical regulatory frameworks—will grow in importance.  

In summary, by synthesizing state-of-the-art methods across different paradigms, this survey highlights the current capabilities and limitations in trajectory prediction and planning. It provides a roadmap for future research, emphasizing the importance of integrating explicit knowledge through symbolic reasoning, hybrid models, or generative approaches. These advancements are crucial for developing the next generation of autonomous driving systems that are both safe and efficient in real-world operations. By advancing knowledge representation techniques and fostering synergistic combinations of different approaches, we can pave the way for more intelligent, trustworthy, and safer autonomous vehicles capable of seamlessly navigating the complexities of real-world driving.

\bibliographystyle{IEEEtran}
\bibliography{reference}
%\printbibliography

\end{document}